# Active Object Manipulation Facilitates Visual Object Learning: An Egocentric Vision Study

Satoshi Tsutsui    Dian Zhi    Md Alimoor Reza    David Crandall    Chen Yu

Indiana University

{stsutsui,dianzhi,mdreza,djcran,chenyu}@indiana.edu

## Abstract

*Inspired by the remarkable ability of the infant visual learning system, a recent study collected first-person images from children to analyze the "training data" that they receive. We conduct a follow-up study that investigates two additional directions. First, given that infants can quickly learn to recognize a new object without much supervision (i.e. few-shot learning), we limit the number of training images. Second, we investigate how children control the supervision signals they receive during learning based on hand manipulation of objects. Our experimental results suggest that supervision with hand manipulation is better than without hands, and the trend is consistent even when a small number of images is available.*

## 1. Introduction

Although some machine learning models for computer vision have achieved superhuman performance when tuned for specific tasks and datasets [4], machine learning generally pales in comparison to the learning system of the human child. Children have the remarkable ability to rapidly learn to recognize visual objects from a few examples, even under very weak supervision. What can we learn from children's learning systems in order to build better visual machine learning? Human learning clearly uses different mechanisms than convolutional neural networks trained with backpropagation, but it also receives very different "training data" — no child learns to recognize objects from millions of photos downloaded from the web, for example. Children also receive very different forms of supervisory signals: while computer vision algorithms usually receive tightly-cropped object boundaries and accurate class labels, children see cluttered environments with only weak supervisory signals to help direct their attention. We are interested in better understanding the properties of the imagery and supervision signals that children's visual systems collect as they go about learning in their everyday lives.

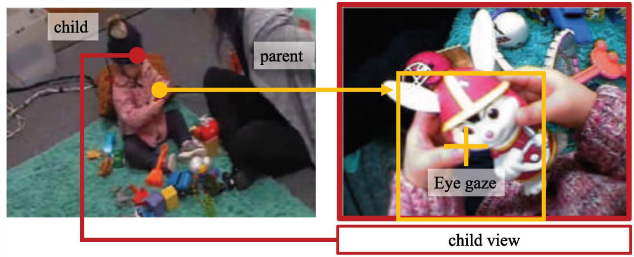

Figure 1. We capture images from a toddler's point of view (red) when a parent teaches the names of toys. We also capture eye gaze and crop the image centered at the gaze point (orange). The crop region corresponds to approximately 30° field of view.

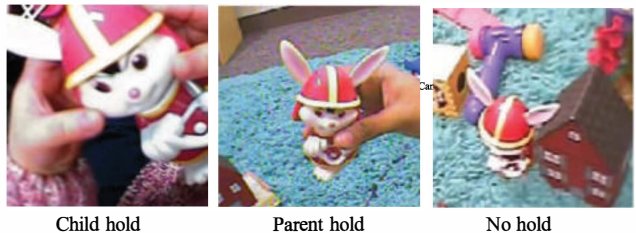

Figure 2. Sample training images. We divide the gaze-centered cropped images into three subsets based on who holds the object. We train classifiers using images from each of the subsets.

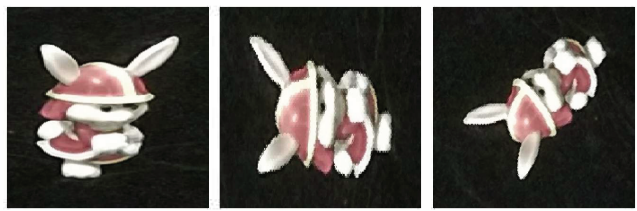

Figure 3. Sample test images, captured with canonical rotations and a clean background. We use test accuracy as a measure of the quality of the training images used to learn the model.

Egocentric cameras provide a way of capturing people's fields of view, allowing us to (approximately) monitor the "input signals" that people's visual systems receive. We use head-mounted cameras on children to capture egocentric imagery as they carry out toy play, which is an important context for real-world learning. We also use an eye gaze tracker to record where the child is attending within the field of view. This technique of applying egocentric computer vision to better understand child visual learning is similar to that proposed by Bambach *et al.* [1]. They collected first-person images from head-mounted cameras on both children and parents playing together in the same environment, and then trained two object recognition models, one using data from the parents and one using data from the children. They then tested the two models on a separate, well-controlled test set of objects photographed against a clean background from canonical views. They demonstrated that object recognition models trained from the child view performed significantly better than those from parents, suggesting that children's visual systems may be receiving higher quality training images.

**Present work**. This paper conducts a follow-up study to Bambach's toddler-inspired learning paper, expanding that work in two major directions. First, we focus on the few-shot training ability of the data collected by toddlers. Instead of the more than 200,000 images collected and used in the original study, we constrain the number of images that we use for training classifiers to mere dozens. Second, we investigate the quality of training data collected by children's egocentric views as a function of types of interactions with objects. In other words, toddlers do not merely observe a scene statically, but are active agents whose visual experiences are controlled both by their own actions and by the actions of the people they interact with (e.g., their parents). Here we investigate the quality of training data during three types of object interactions: (1) when a toddler looks at an object held by a parent, meaning that the parent is supervising the toddler's visual learning; (2) when a toddler looks at and holds a target object, which means they are actively self-supervising the training; and (3) when a toddler just looks at a target object left on the floor, which is the weakest version of supervision. A sample from each type is shown in Figure 2.

We conduct a controlled experiment to train a simple image classifier from the three types of supervising signals with a relatively small number of images. The accuracy of the resulting classifier indicates that supervision without hands is always worse than with parents' or children's hands. Within hand based supervision, children's own signal is equally good as parent's supervision.

**Contribution**. In summary, this paper has the following two contributions for toddler-inspired visual learning.

1. Inspired from toddlers' few-shot learning ability to recognize objects, we limit the number of images for training and investigate the performance in few-shot situation.

2. We divide the toddlers' supervision signal into several types and show that hand based active object manipulation provides a higher quality of supervision signals.

## 2. Related Work

Toddler-inspired visual learning [1] lies in the intersection of computer vision (including first-person vision), machine learning, and psychology. We refer to the original paper for broadly-defined related literature in these fields, and here we discuss the difference between our few-shot experiment and what is generally called few-shot learning.

In contrast to classical few-shot learning [3], recent few-shot work [6–8] adapts a meta-learning problem with disjoint object categories in training and testing. In the training (or formally meta-training) stage, the model learns a better representation for learning with a few examples per class. In the meta-test stage, the model has to classify unseen images of unseen classes provided a few examples per class. For example, the simplest way to do this is a nearest neighbor classifier using the learned representation. Our way of few-shot classification is *not* this meta-learning setup, and meta-learning is not our interest. We are interested in the training data that children collect for quickly adapting new classes, so we only do what they call meta-testing stage, using a simple classifier with a nearest neighbor approach.

## 3. Experiments

### 3.1. Dataset and Methodology

We use the Indiana toddler-learning dataset [1] with their evaluation methodology. The dataset contains first-person videos of parent and children in a scenario where toddlers learn about objects (toys) and their names guided by parents (see Figure 1). The total number of parent-child pairs is 26, and the total number of toys is 24. The toddlers are from 15 to 24 months old. Each frame has eye-gaze fixations as well as bounding boxes of the toys, which in combination indicates which toy (if any) is attended in each frame. When a toddler attends an object, an image of the object is cropped centered around the gaze, and regarded as a labeled image for training a classifier in a supervised manner. The classifier is tested with a separate set of images [1] where each object is systematically photographed from 128 different views and distances, totaling 3,072 images. We show sample test images in Figure 3, and show more images in the Appendix. The test accuracy is considered as an indicator of the quality of the supervision signal.

We investigate how toddlers' visual systems seek a supervision signal, so we only use toddler-centric videos. We manually label the "holding status" of each attended object on a frame by frame basis, dividing frames into three categories: (1) **Parent-hold supervision**, where the toddler looks at the object held by its parent; in this case, the visual system is supervised from externally controlled signals; (2) **Child-hold self-supervision**, where the toddler looks at the object held by him- or herself, corresponding to the toddler self-supervising the training; and (3) **No-hold self-supervision**, when the toddler looks at an object that is not held by anybody, which is the weakest type of supervision. We show a sample from each set in Figure 2. To divide the frames into these three groups, we manually annotated a subset of 249,287 frames, where 70,869 are parent-hold, 99,190 are child-hold, and 79,228 are no-hold.

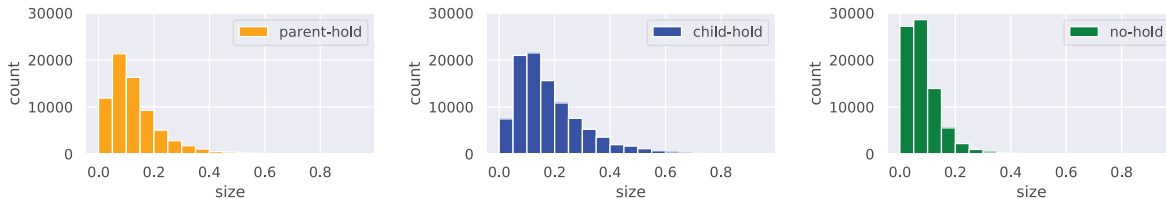

Figure 4. The size distribution for child-hold, parent-hold, and no-hold. The size is defined as proportion of the object bounding box to the child's field of view.

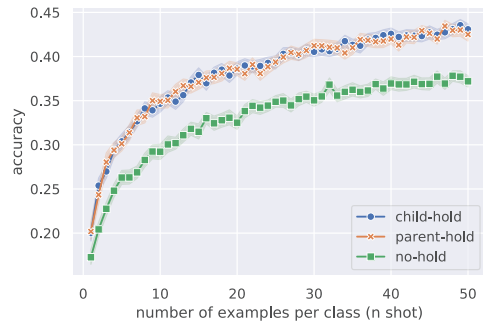

Figure 5. Few-shot classification accuracy. The marker is mean accuracy over 100 trials and the band is 95% confidence interval.

In order to simulate the toddlers' view, we crop the image because the camera field of view (70°) is larger than a human's. Instead of Bambach *et al.* [1]'s approach to simulate various fields of view with Gaussian blur acuity effects, we simply crop a square corresponding to the highest acuity region (30°), centering at the gaze fixation point.

We are interested in the case where a system is trained on a relatively small amount of data — i.e., few-shot situations. We train a classifier with a limited number of training examples per object, ranging from one to fifty. As our focus is neither the algorithm side nor improving the performance of a classifier, we adopt a very simple baseline for few-shot learning: a nearest neighbor classifier from an ImageNet-pretrained convolutional neural network (VGG16 [5]). We know that ImageNet pretraining is likely not what toddlers' brains are doing. On one hand, it makes sense to use some pretrained representation because toddlers (of 19 months old on average in the dataset) have trained their visual sys-tem with the equivalent of millions of images. On the other hand, the quality of images from ImageNet is very different from toddlers' experiences. We leave this question for future work and just approximate their internal visual representation with the ImageNet features, as in the original study [1]. In practice, we need some feature vectors even though we merely use the simplest classification baseline for few-shot learning. We then use the accuracy of the classifiers on the canonical test as an evaluation metric.

### 3.2. Results and Discussion

We show the classification accuracy in Figure 5. To our surprise, the parent-hold case and child-hold case have no significant difference. This means parental supervision to the toddler's learning system is as good as their own self-supervision. On the other hand, the supervision signal without hand-holding is significantly worse than the hand based one. The observed trend is consistent in terms of the number of training images.

In order to investigate which visual characteristics distinguish the holding image set from the gaze-only set, we follow Bambach *et al.* [1] and visualize the object size distribution of each set in Figure 4 where the size means the proportion of object bounding box to the child's field of view. It clearly shows that the size tends to be smaller when objects are not being held, which we believe is one reason for the performance loss. However, it is also true that the child-hold set's distribution has a longer tail than parent-hold, thus having more larger objects on average, but the performance is almost the same. This suggests that parent-hold images may have some undiscovered characteristics that make them suitable to train a classifier and compensate for the size gap from the child-hold images.

## 4. Conclusion

We conducted few-shot classification experiments of toddler-inspired visual learning in order to investigate the effectiveness of different supervision signals that children receive in their daily life for training their internal visual system. Our experimental results suggest that hand-based supervision is more effective than weaker supervision. Within hand based supervision, our results indicate that parent-hand manipulation and children's own hand manipulation have no significant difference.

## Acknowledgment

We would like to thank Yi Li for drawing Figure 1, Shujon Naha and Andrei Amatuni for useful discussions, and Sven Bambach for helping prepare the data. This work was supported by the National Science Foundation (CAREER IIS-1253549) and the National Institutes of Health (R01 HD074601, R01 HD093792), as well as the IU Office of the Vice Provost for Research, the College of Arts and Sciences, and the School of Informatics, Computing, and Engineering through the Emerging Areas of Research Project "Learning: Brains, Machines, and Children."

## Appendix: Test Images

We use the same test images from previous work [1]. We have 24 different toys and show them in Figure 6. We systematically capture the toys from various views, sizes, and rotations, and show examples in Figure 7.

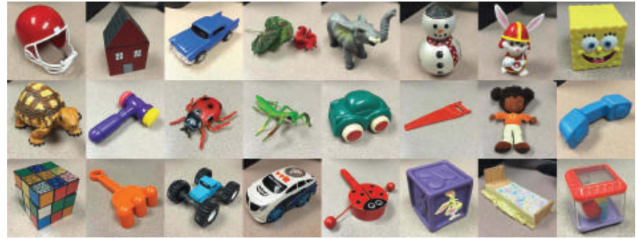

Figure 6. The 24 toys used in our study (copied from [2]). Note that we do not use these images, and use black background for test, and cluttered room background for training.

Figure 7. All test images of one of our toys. We capture each toy from eight different point of views in each 45 angle rotation around its vertical axis. We computationally transform each view for two sizes and eight rotations (0, 45, 90, ..., 315). We then resize the images into squares as inputs for CNNs. Bambach *et al*. [2] write more details about the test image capturing procedure.

4